\pdfoutput=1

\documentclass[11pt]{article}

\usepackage[final]{acl}

\usepackage{enumitem}

\usepackage{hyperref}
\usepackage[figuresright]{rotating}

\usepackage{multirow}
\usepackage{adjustbox}
\usepackage{booktabs}
\usepackage{amsthm,amsmath,amssymb}
 
\usepackage{mathrsfs}

\usepackage{times}
\usepackage{latexsym}

\usepackage[T1]{fontenc}

\usepackage[utf8]{inputenc}

\usepackage{microtype}

\usepackage{inconsolata}
\usepackage{amsmath}
\usepackage{amsfonts}
\usepackage{url}
\usepackage{graphicx}

\usepackage{multirow}
\usepackage{multicol}
\usepackage{booktabs}
\usepackage{arydshln}
\usepackage{color}
\usepackage{makecell}

\usepackage{graphicx}
\usepackage{comment}
\usepackage{booktabs}

\usepackage{colortbl}
\definecolor{mygray}{gray}{.9}

\title{Stable Knowledge Editing in  Large Language Models}

\author{Zihao Wei$^{1,2}$\quad
Liang Pang$^{1}$\thanks{\ \ Corresponding author}\quad
Hanxing Ding$^{1,2}$\quad
Jingcheng Deng$^{1,2}$\quad
\textbf{Huawei Shen}$^{1,2}$\quad
\textbf{Xueqi Cheng}$^{1,2}$\\
 $^{1}$ Institute of Computing Technology, Chinese Academy of Sciences \\
 $^{2}$ University of Chinese Academy of Sciences \\
 \texttt{\{weizihao22z, pangliang, dinghanxing18s, dengjingcheng23s, shenhuawei, cxq\}@ict.ac.cn}
}

\begin{document}
\maketitle
\begin{abstract}
Efficient knowledge editing of large language models is crucial for replacing obsolete information or incorporating specialized knowledge on a large scale. 
However, previous methods implicitly assume that knowledge is localized and isolated within the model, an assumption that oversimplifies the interconnected nature of model knowledge.
The premise of localization results in an incomplete knowledge editing, whereas an isolated assumption may impair both other knowledge and general abilities.
It introduces instability to the performance of the knowledge editing method.
To transcend these assumptions, we introduce StableKE, a method adopts a novel perspective based on knowledge augmentation rather than knowledge localization.
To overcome the expense of human labeling, StableKE integrates two automated knowledge augmentation strategies: Semantic Paraphrase Enhancement strategy, which diversifies knowledge descriptions to facilitate the teaching of new information to the model, and Contextual Description Enrichment strategy, expanding the surrounding knowledge to prevent the forgetting of related information.
StableKE surpasses other knowledge editing methods, demonstrating stability both edited knowledge and multi-hop knowledge, while also preserving unrelated knowledge and general abilities. 
Moreover, StableKE can edit knowledge on ChatGPT.
\end{abstract}

\section{Introduction}

Extensive research has consistently shown that large language models (LLMs) possess the capability to harness the vast reservoir of knowledge stored within their parameters for various reasoning tasks. However, this ability comes with inherent risks, including the potential for these models to inadvertently absorb obsolete or incorrect information~\cite{DBLP:journals/jmlr/RaffelSRLNMZLL20,DBLP:conf/nips/BrownMRSKDNSSAA20,DBLP:conf/nips/Ouyang0JAWMZASR22,DBLP:journals/corr/abs-2302-13971,DBLP:journals/corr/abs-2307-09288,DBLP:journals/corr/abs-2303-18223}. Hence, the concept of knowledge editing in LLMs is introduced to address the timely updating of obsolete information and the integration of specialized knowledge on a large scale~\cite{DBLP:conf/iclr/SinitsinPPPB20,DBLP:journals/corr/abs-2104-10350,DBLP:conf/nips/LazaridouKGALTG21,dhingra-etal-2022-time}.

\begin{figure}[t]
  \centering
  \includegraphics[width=\columnwidth]{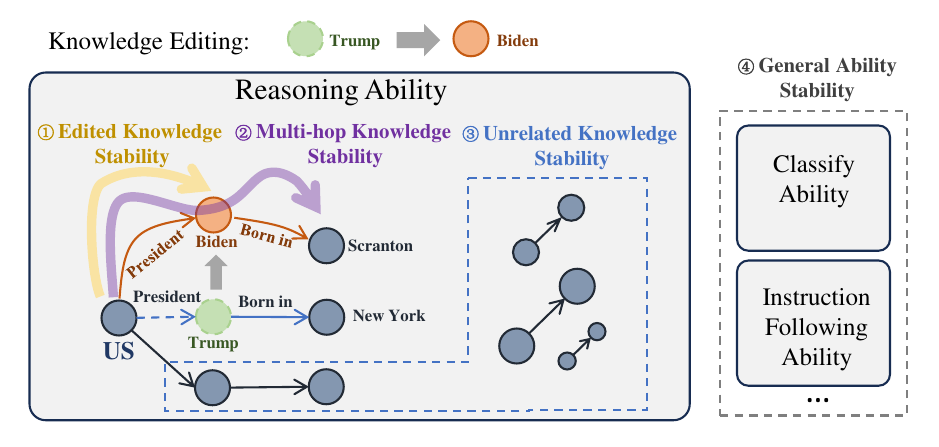}
  \caption{To enhance the evaluation of a knowledge editing method, we propose assessing it across four dimensions of stability. (1) Edited Knowledge Stability reflects the performance of one-hop knowledge editing, focusing on the consistency and accuracy of edited knowledge. (2) Multi-hop Knowledge Stability evaluates how well the edited knowledge integrates with existing knowledge across multiple steps. (3) Unrelated Knowledge Stability and (4) General Ability Stability, ensures that unrelated knowledge remains unchanged and maintain the overall capabilities of the model despite the editing process.}
  \label{fig:motivation}
  \vspace{-0.5cm}
\end{figure}

Previous knowledge editing methods include locate-then-edit~\cite{dai-etal-2022-knowledge,DBLP:conf/nips/MengBAB22,DBLP:conf/iclr/MengSABB23}, memory-based models~\cite{DBLP:conf/icml/MitchellLBMF22,DBLP:conf/iclr/HuangSZZR023,DBLP:conf/emnlp/DongDSXSL22}, and meta-learning~\cite{de-cao-etal-2021-editing,DBLP:conf/iclr/MitchellLBFM22} have two implicit assumptions to the knowledge, (1) knowledge is encapsulated in localized parameters, such as parameters in LLMs or external memory modules, (2) knowledge is independent of each other as well as isolated from the general capabilities of LLMs.
However, the study conducted by~\cite{DBLP:journals/corr/ABS-2311-09053} challenges the initial assumption by demonstrating that knowledge in LLMs is not solely confined to the MLP layers but also resides within the attention layers. Consequently, adjusting the weights where knowledge is stored has been shown to be ineffective~\cite{DBLP:journals/corr/abs-2301-04213}.
The second assumption overlooks the intricate interconnections between knowledge, which can result in catastrophic forgetting of unrelated knowledge and impact the general abilities of LLMs~\cite{wang-etal-2022-finding-skill}.

These problems can be summarized as four aspects of knowledge editing stability to evaluate the effectiveness of knowledge editing methods. 
(1)~\textbf{Editing Knowledge Stability}: For the knowledge currently being edited, ensure consistency in describing the semantic variance of the questions related to the edited knowledge. For example, change the question from `how many' to `what is the number'.
(2)~\textbf{Multi-hop Knowledge Stability}: For the subsequent multi-hop knowledge, we expect the model to propagate these knowledge changes to the associated questions. For instance, illustrated in Figure~\ref{fig:motivation}, transitioning from the Trump to Biden presidency necessitates updating the birthplace of the president.
(3)~\textbf{Unrelated Knowledge Stability}: For other unrelated knowledge, we prioritize preserving the stability of unrelated knowledge, ensuring it remains unaffected by the editing process. 
(4)~\textbf{General Capability Stability}: For general capabilities, such as classify and instruction following abilities, efforts are made to minimize their impact on the overall capabilities of the model.
Furthermore, to ensure necessary practice, the editing process must exhibit stability on a large scale, encompassing batch knowledge editing and sequential knowledge editing.

To avoid the instability inherent in traditional methods, we advocate for a novel strategy focused on knowledge augmentation rather than knowledge localization. This approach necessitates access to a vast and diverse dataset related to both edited knowledge and unrelated information. To alleviate the burden of data labeling from human labor and harness the paraphrasing capabilities of advanced LLMs, we propose StableKE, which incorporates two automated knowledge augmentation strategies: Semantic Paraphrase Enhancement (SPE) and Contextual Description Enrichment (CDE).
The SPE strategy, which diversifies knowledge descriptions to facilitate the teaching of new information to the model, akin to the diverse learning experiences in human cognition~\cite{sanger2020diversity,auerbach2012teaching,stern2017individual}. 
The CDE strategy, through the development of a comprehensive dataset of descriptive texts, bolsters the capacity of the model to retain relevant information, effectively circumventing issues of knowledge forgetting.

To verify stabilities of knowledge editing methods on a large scale and mitigate the limitations of existing knowledge editing datasets (discuss in detail in \S\ref{Knowledge Editing Dataset}), we introduce a tree-structured multi-hop knowledge editing dataset and metrics for stabilities, namely KEBench. This benchmark is crafted to support intricate, real-world multitasking and hierarchical reasoning evaluations, offering a nuanced assessment of the impact of editing on model reasoning capability.

The StableKE method showcases remarkable performance across four types of stabilities compared to other baseline knowledge editing methods. It maintains consistently strong performance on existing knowledge and subsequent reasoning steps while preserving other knowledge and model abilities unchanged.
Particularly in large-scale settings such as batch editing and sequential editing, its performance remains stable even as some state-of-the-art methods approach model collapse.
Notably, StableKE has also achieved high accuracy in editing tasks using the state-of-the-art GPT-3.5-turbo through the ChatGPT fine-tuning API~\footnote{\url{https://platform.openai.com/docs/api-reference}}.

In summary, our contributions are as follows:

\begin{itemize}[itemsep=2pt,topsep=0pt,parsep=0pt]
    \item Identifying two key assumptions in existing knowledge editing methods and outline how these assumptions lead to four types of instability issues in knowledge editing.
    \item Introducing StableKE, a novel approach that underscores the pivotal role of data in refining knowledge editing.
    \item Developing a comprehensive tree-structured dataset tailored for evaluating knowledge editing methods against critical stability criteria.
\end{itemize}
\section{Related Work}
We introduce recent knowledge editing methods and datasets in this section.

\subsection{Knowledge Editing Methods}

Current knowledge editing methods by modifying the model parameters of LLMs can be divided into three main paradigms based on where knowledge is stored and the learning approach employed: locate-then-edit, memory-based models, and meta-learning. 

\textbf{Locate-Then-Edit}: This paradigm first identifies a subset of parameters in the model that are related to the edited knowledge, and then updates them to perform knowledge editing. For instance, \citet{dai-etal-2022-knowledge} manipulates `knowledge neurons' (KN) in pretrained transformers to update facts. Similarly, \citet{DBLP:conf/nips/MengBAB22} introduces a method for editing factual associations in LLMs by modifying key feedforward weights using Rank-One Model Editing (ROME). However, ROME and KN can only modify one piece of knowledge at a time. To this end, \citet{DBLP:conf/iclr/MengSABB23} expanded the settings of ROME and built MEMIT so that it can change a batch of knowledge at 
once.

\textbf{Memory-based Model}: This paradigm facilitates editing through the integration of a small auxiliary model or the addition of extra parameters within the MLP layer, while keeping the parameters of original model fixed. SERAC, which modifies knowledge by optimizing a counterfactual model~\cite{DBLP:conf/icml/MitchellLBMF22}, and T-Patcher, which achieves knowledge editing by incorporating a small number of trainable neuron patches into the MLP layer~\cite{DBLP:conf/iclr/HuangSZZR023}. Furthermore, CALINET utilizes the properties of MLP layers to directly calibrate factual knowledge in LLMs~\cite{DBLP:conf/emnlp/DongDSXSL22}.

\textbf{Meta-learning}: This paradigm employs a hypernetwork designed to master the alterations required for the manipulation of knowledge in the MLP layers of models. such as KnowledgeEditor~\cite{de-cao-etal-2021-editing}, leverage hypernetworks for efficient language model edits. MEND~\cite{DBLP:conf/iclr/MitchellLBFM22}, introducing auxiliary networks, allows scalable edits by decomposing gradients. 

These approaches are based on two basic assumptions: knowledge is localized in parameters and isolated with each other. 

\subsection{Knowledge Editing Dataset}
\label{Knowledge Editing Dataset}
In exploring the effectiveness of knowledge editing in LLMs, research has demonstrated datasets to evaluate knowledge editing methods.
RIPPLEEDITS~\cite{DBLP:journals/corr/abs-2307-12976}, with 5,000 factual editing cases, serves as a diagnostic benchmark, aiming to capture the cascading effects of knowledge edits. 
MQuAKE~\cite{DBLP:journals/corr/abs-2305-14795} focuses on multi-hop questions, evaluates the impact of edits on complex knowledge chains.
However, existing datasets are unable to comprehensively evaluate all four types of stability. For instance, (1) they often lack tests involving multiple relations related to a single knowledge edit, resulting in significant performance variance; (2) they fail to verify the unchanged relations of both the source and target edited objects, making it hard to ascertain whether other knowledge remains preserved in LLMs; (3) the overall capabilities of the LLMs are often overlooked in these datasets. Therefore, we propose a new benchmark, KEBench, to address these challenges and provide a thorough evaluation of knowledge editing methods. 
\section{KEBench Benchmark}
To address the shortcomings of previous knowledge editing datasets, we present KEBench, a comprehensive benchmark for evaluating knowledge editing methods. 

\subsection{Task Formalization}

\begin{figure}
  \centering
  \includegraphics[width=\columnwidth]{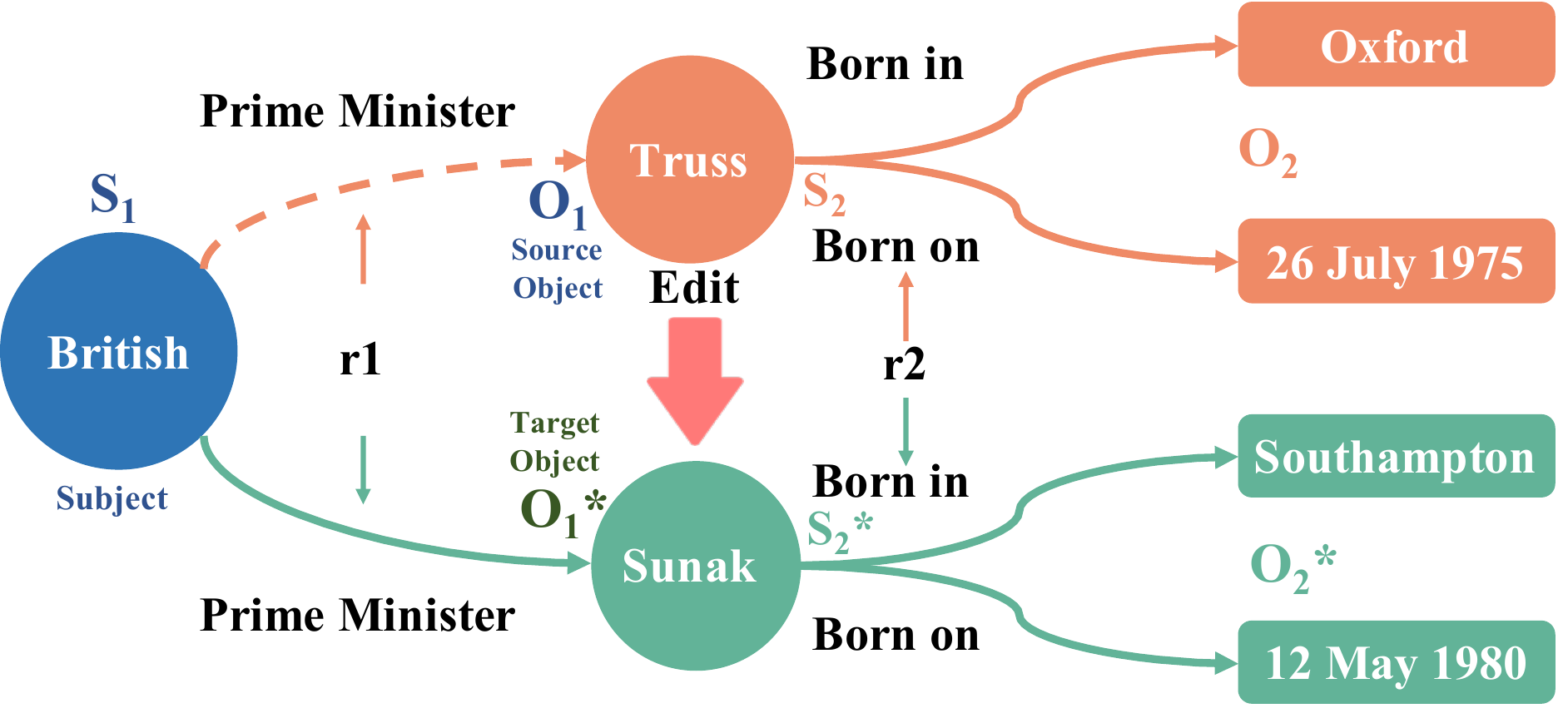}
  \caption{An example of our KEBench.}
  \label{fig:dataset}
\end{figure}

Following previous work ~\cite{DBLP:conf/emnlp/YaoWT0LDC023,DBLP:journals/corr/abs-2308-07269,gu2023pokemqa}, we define fact triples as $t = (s, r, o)$, where the subject $s$ and the object $o$ are entities, and $r$ is a relation, for example \textit{(British, Prime Minister, Truss)} in Figure~\ref{fig:dataset}. 
To precisely reflect the transition in the \textit{British Prime Minister} position from \textit{Truss} to \textit{Sunak} on Oct. 25, 2022, we adjusted fact triple to $t^* = (s, r, o^*)$, where the target object $o^*$ \textit{Sunak} replaces the source object $o$ \textit{Truss}.
Fundamental to the process of knowledge editing, this process refinement is succinctly captured by $e = (s, r, o \rightarrow o^*)$, marking the key transition of our computational model from its base state $f_\theta$ to an enhanced iteration $f_{\theta_e}$.

\subsection{Data Collection}

Our dataset comprises an extensive collection of structured two-hop triplets. Firstly, we collect a common set $\mathcal{R}$ of relationships based on previous work \cite{petroni-etal-2019-language, DBLP:conf/nips/MengBAB22}, where $|\mathcal{R}|=37$. To generate first-hop data, we selected fact triples $t_1 = (s_1,r_1,o_1)$ for each relation $r_1 \in \mathcal{R}$ based on Wikidata. Then, we use the template to generate first-hop questions $Q(t_1)$ and answers $A(t_1)$. For each fact triple, we also need to construct a set of knowledge editors $e=(s_1,r_1,o_1\rightarrow o_1^*)$ to obtain $t_1^* = (s_1,r_1,o_1^*)$. Correspondingly, we can obtain $A(t_1^*)$. Since the subject $s_1$ and relation $r_1$ are the same, $Q(t_1^*)$ equals $Q(t_1)$. To obtain second-hop data, we collected fact triples that share a common relationship $r_2 \in \mathcal{R}^* \subseteq \mathcal{R}$ ("born in" and "born on" in ~\ref{fig:dataset}) for entities $o_1$ and $o_1^*$, denoted as $t_2 = (s_2, r_2, o_2)$ and $t_2^* = (s_2^*, r_2, o_2^*)$, where $s_2$ and $s_2^*$ represent $o_1$ and $o_1^*$ respectively. Likewise, for each $t_2$ (or $t_2^*$) we can get Q($t_2$) (or Q($t_2^*$)) and A($t_2$) (or A($t_2^*$)). 

Utilizing the multi-hop question generation method presented in \citet{DBLP:journals/corr/abs-2305-14795}, we generate multi-hop questions $Q_h(t_1^*,t_2^*)$ that delve into the interconnections between numerous $o_2^*$ entities and a single distinguished $o_1^*$ entity. Subsequent experiments prove that even a two-hop problem poses a great challenge to existing knowledge editing methods, so for convenience, we do not construct more complex problems (larger than two hops). Some rules were adopted to filter the fact triplets (see Appendix~\ref{app:Sampling} for details). Our dataset, featuring 1000 knowledge editing triples and 2798 multi-hop questions, establishes a comprehensive benchmark for examining the four critical stability principles within knowledge editing.

\subsection{Evaluation Metrics}
\label{sec:metric}

In order to test the performance of Edited Knowledge Stability, Multi-hop Knowledge Stability, Unrelated Knowledge Stability and General Ability Stability. We assessed the above three principles from seven perspectives.

\paragraph{Edited Knowledge Stability} The most direct way is to measure the average accuracy of the first-hop data.
We adopt two metrics for this stability, one for direct questions, one for paraphrased questions.
For direct questions, we express it as 1Hop Question Accuracy (1Hop-Acc), which also noted as Reliability in the work \citet{DBLP:conf/emnlp/YaoWT0LDC023},
\begin{equation}
    \small \text{1Hop-Acc}=\mathbb{E}(\mathbb{I}_{A(t_1^*)}(f_{\theta_e}(Q(t_1^*)))),
    \label{eq1}
\end{equation}
where $\mathbb{I}_A(x)$ is the indicator function, which takes 1 if $x$ belongs to set $A$. 

For paraphrased questions, we use ChatGPT to generate the original question into a paraphrase sentence with the same semantics and different expressions, and use the paraphrase question to test the model knowledge. We express it as 1Hop Question Accuracy (Paraphrased) and abbreviate it as Para-1Hop-Acc. It is similar with the Generalization in \cite{DBLP:conf/emnlp/YaoWT0LDC023}:
\begin{equation}
    \small \text{Para-1Hop-Acc}=\mathbb{E}(\mathbb{I}_{A(t_1^*)}(f_{\theta_e}(Q_p(t_1^*)))),
    \label{eq2}
\end{equation}
where $Q_p(t_1^*)$ is $Q(t_1^*)$ rephrased using ChatGPT.

\paragraph{Multi-hop Knowledge Stability} We define three metrics to evaluate stability, which are not available in other benchmarks. Firstly, in order to ensure that the model can answer multi-hop questions with knowledge, we test questions $Q(t_1^*)$ and $Q(t_2^*)$, and measure the average accuracy of $Q(t_2^*)$ while $Q(t_1^*)$ is correct. This primarily assesses the capacity of the model for reasoning. We express it as Decomposed 2Hop Question Accuracy (Decom-2Hop-Acc):
\begin{equation}
\small
\begin{aligned}
    &\text{Decom-2Hop-Acc}=\\
    &\mathbb{E}(\mathbb{I}_{A(t_1^*)}(f_{\theta_e}(Q(t_1^*)))*\mathbb{E}(\mathbb{I}_{A(t_2^*)}(f_{\theta_e}(Q(t_2^*))))).
    \label{eq3}
\end{aligned}
\end{equation}

Then we directly test the accuracy of the model in answering multi-hop questions $Q_h(t_1^*,t_2^*)$. This not only tests the reasoning ability of  the model, but also tests the knowledge contained in the model. We express it as Composed 2Hop Question Accuracy (Com-2Hop-Acc): 
\begin{equation}
    \small \text{Com-2Hop-Acc}=\mathbb{E}(\mathbb{I}_{A(t_1^*,t_2^*)}(f_{\theta_e}(Q_h(t_1^*,t_2^*)))).
    \label{eq4}
\end{equation}

In addition, we also examine the instruction of the model compliance capabilities and CoT capabilities. We build $Q_h^{\mathrm{CoT}}(t_1^*,t_2^*)$ to guide the model to follow the instructions by adding "Please provide a multi-hop explanation for the next question" to the input instructions. We measure the average accuracy of the answers of the edited model and express it as Composed 2Hop Question Accuracy (with CoT), abbreviated as Com-2Hop-Acc (CoT):
\begin{equation}
\small 
\begin{aligned}
    \text{Com-2Hop}&\text{-Acc(CoT)}=\\
    &\mathbb{E}(\mathbb{I}_{A(t_1^*,t_2^*)}(f_{\theta_e}(Q_h^{\mathrm{CoT}}(t_1^*,t_2^*)))).
    \label{eq5}
\end{aligned}
\end{equation}

\paragraph{Unrelated Knowledge Stability} These metrics are also called Locality or Specificity \cite{DBLP:conf/emnlp/YaoWT0LDC023}. Different from other baseline methods that randomly sample knowledge to evaluate, in order to downgrade the evaluation variance, we test whether the edited model can answer relevant knowledge of source entity $o_1$ and target entity $o_1^*$. In other words, we separately test the accuracy of the edited model in answering the second-hop questions $Q(t_2)$ and $Q(t_2^*)$. Expressed by Unrelated Question on Source Entity Accuracy (Src-Acc) and Unrealted Question on Target Entity Accuracy (Tgt-Acc):
\begin{equation}
\small 
    \text{Src-Acc}=\mathbb{E}(\mathbb{I}_{A(t_2)}(f_{\theta_e}(Q(t_2)))),
    \label{eq6}
\end{equation}
\begin{equation}
\small 
    \text{Tgt-Acc}=\mathbb{E}(\mathbb{I}_{A(t_2^*)}(f_{\theta_e}(Q(t_2^*)))).
    \label{eq6}
\end{equation}

\paragraph{General Capability Stability} We directly report the average score of the edited model on the MMLU benchmark~\cite{DBLP:conf/iclr/HendrycksBBZMSS21}.
\section{StableKE Method}
Recognizing various limitations of previous knowledge editing methods that relied on two controversial assumptions, we pay more attention to knowledge augmentation strategy, which is used to improve the stability of model knowledge editing without relying on those assumptions.

\subsection{Overview}
StableKE method leverages semantic paraphrase enhancement (SPE) and contextual description enrichment (CDE) to enrich semantic and contextual understanding of the model, thereby enhancing its resilience and adaptability.

\subsection{Semantic Paraphrase Enhancement}
We introduce an innovative methodology, semantic paraphrase enhancement (SPE), aimed at enriching comprehension within the model by exposing it to multiple semantic representations of the same concept. This approach is akin to multifaceted learning in human education\cite{sanger2020diversity,auerbach2012teaching,stern2017individual}, where diverse explanations and perspectives solidify understanding. Utilizing advanced generative models, we produce $K_{spe}$ distinct textual variations $A_i^{p}(t_1^*)$ for answers to every first-hop question $A(t_1^*)$:
\begin{equation}
\small 
    A_i^{p}(t_1^*) = \mathrm{ChatGPT}(P_{spe}(A(t_1^*))), i \in [1,K_{spe}],
    \label{eq7}
\end{equation}
where $P_{spe}$ stands for SPE prompt template (see Appendix~\ref{app:c}  for details) and $\mathrm{ChatGPT}$ represents the output of ChatGPT. This ensures varied semantic conveyance while maintaining a balance between creativity and coherence, achieved through a calibrated generation temperature of 0.7. We proceed by pairing each question $Q(t_1^*)$ with its corresponding answer $A_i^{p}(t_1^*)$, forming question-answer pairs ($Q(t_1^*)$, $A_i^{p}(t_1^*)$). These data are then fed into the model for training.

\subsection{Contextual Description Enrichment}

The contextual description enrichment (CDE) is developed through meticulous selection and curation of descriptive texts, focusing on both the original $o_1$ and modified entities $o_1^*$. For each entity $o$, we search related terms from Wikidata and form a document $\mathrm{doc}(o)$. Then, in order to ensure semantic coherence and richness, we use ChatGPT to rewrite it to obtain $K_{cde}$ different expressions:
\begin{equation}
\small 
    \mathrm{doc}_i^p(o) = \mathrm{ChatGPT}(P_{cde}(\mathrm{doc}(o))), i \in [1,K_{cde}],
    \label{eq8}
\end{equation}
where $\mathrm{doc}_i^p(o)$ represents the $i$th document rewritten using ChatGPT about entity $o$ and $P_{cde}$ stands for CDE prompt template (see Appendix~\ref{app:c}  for details). The CDE is integral to the Post-Pretrain phase, which can significantly enhance the ability of the model to retain and integrate relevant information, effectively avoiding the problem of knowledge forgetting. We combine 
documents $\mathrm{doc}_i^p(o)$ and corresponding instructions (such as 'Please describe the sunak.') to construct data to allow the model to perform instruction fine-tuning.

\subsection{Model Training}
In addition to the original question-answer pairs $(Q(t_1^*),A(t_1^*))$, we also employ data generated by SPE and CDE for supervised instruction fine-tuning. The mixing ratio of SPE and CDE generated data is 3:5. Unlike typical fine-tuning, instruction fine-tuning zeroes out the token loss within the questions during the training phase, allowing backpropagation to focus exclusively on factual descriptions. We implemented a cosine learning rate schedule with an initial learning rate of $2\times 10^{-5}$.
\section{Experiments}
In this section, we present our experimental setups and meticulously analyze the results.

\begin{table*}[t]
\centering
\resizebox{1\linewidth}{!}{
\begin{tabular}{lcccccccc}
\toprule
\multirow{2}{*}{\textbf{Method}} & \multicolumn{2}{c}{\textbf{Edited}} & \multicolumn{3}{c}{\textbf{Multi-Hop}} & \multicolumn{2}{c}{\textbf{Unrelated}} & \multicolumn{1}{c}{\textbf{General}}\\
\cmidrule(r){2-3} \cmidrule(r){4-6} \cmidrule(r){7-8} \cmidrule(r){9-9}
 & \multicolumn{1}{c}{\textbf{\small{1Hop-Acc}}} & \multicolumn{1}{c}{\textbf{\small{Para-1Hop-Acc}}} & \multicolumn{1}{c}{\textbf{\small{Decom-2Hop-Acc}}} & \multicolumn{1}{c}{\textbf{\small{Com-2Hop-Acc}}} & \multicolumn{1}{c}{\textbf{\small{Com-2Hop-Acc (CoT)}}} & \multicolumn{1}{c}{\textbf{\small{Src-Acc}}} & \multicolumn{1}{c}{\textbf{\small{Tgt-Acc}}} & \multicolumn{1}{c}{\textbf{\small{MMLU}}} \\ 
\midrule
Finetune & 8.50 & 8.80 & 0.36 & 1.61 & 3.43 & 4.75 & 5.75 & 41.44 \\
Prompt & 18.00 & 17.40 & 21.05 & 20.37 & 25.34 & 69.76 & 54.22 & -\\
SERAC
& 8.40  & 3.20  & 0.00  & 0.50  & 0.43  & 0.79  & 0.18 & -\\
MEND
& 0.00  & 0.00  & 0.00  & 0.00  & 0.00  & 0.00  & 0.00  & 23.14 \\
ROME
& 6.10 & 4.90 & 0.00 & 0.21 & 0.18 & 0.00 & 0.04  & 23.30 \\
MEMIT
& 41.40 & 44.00 & 13.27 & 10.36 & 5.40 & 24.91 & 18.98 & 25.23 \\
\midrule
StableKE+LoRA & 89.10 & 49.50 & 32.27 & 21.55 & 18.76 & 41.92 & 38.17 &40.48 \\
StableKE & \textbf{89.40} & \textbf{83.80} & \textbf{77.09} & \textbf{28.84} & \textbf{31.06} & \textbf{87.03} & \textbf{83.81} & \textbf{42.49}\\
\bottomrule
\end{tabular}
}
\caption{Performance comparison between StableKE and other methods across edited knowledge, multi-hop knowledge, unrelated knowledge, and general ability stability.}
\label{tab:main}
\end{table*}

\subsection{Experimental Setups}
In this paper, our focus lies on practical large-scale knowledge editing, which can be divided into two factors: batch editing and sequential editing. Batch editing involves modifying a significant amount of knowledge collectively in one model update, with the number of examples in one batch denoted as $N_{batch}$. Sequential editing, on the other hand, entails adjusting knowledge batch by batch through multiple iterations, with the number of iterations denoted as $N_{seq}$. Consequently, the total number of edited knowledge instances of the model is calculated as $N_{batch}\times N_{seq}$.

Then, we pose the following four research questions and answer them through corresponding experiments.

\begin{itemize}[itemsep=2pt,topsep=0pt,parsep=0pt]

\item RQ1: How is the stability of knowledge editing when applied in large scale?
\item RQ2: Does the stability of knowledge editing change when applied in batch or a sequential manner?
\item RQ3: How does the quantity of semantic paraphrases affect the performance of StableKE?
\item RQ4: Is StableKE versatile enough for various fine-tuning approaches?
\end{itemize}

The baselines setups and implementation details are described in Appendix~\ref{sec:implementation}.

\subsection{Evaluating Stability in Large Scale}

RQ1 focuses on comparing the abilities of different knowledge editing methods conditional on the same amount of edited knowledge, e.g. $N_{batch}\times N_{seq}=1000$. In this setting, we choose the best batch size $N_{batch}$ and number of iterations $N_{seq}$ for each methods and evaluate the stability performance of large-scale knowledge editing through four aspects mentioned in \S\ref{sec:metric}. 

From the results presented in Table~\ref{tab:main}, it is evident that StableKE exhibits a substantial performance advantage over previous methods, achieving accuracy of $89.40\%$ and $83.80\%$ in the two edited knowledge metrics. These results underscore the superiority of our method in knowledge editing. Additionally, it is apparent that the model parameters of MEND, ROME, and SERAC experience significant degradation under the large-scale setting, resulting in complete failure in knowledge editing tasks.

In the `Multi-Hop' category of Table~\ref{tab:main}, our model demonstrates exceptional performance in the Com-2Hop and Com-2Hop (CoT) metrics, achieving accuracy of $28.84\%$ and $31.06\%$, respectively. It is worth noting that while the Prompt method benefits from reduced difficulty in multi-hop reasoning by explicitly providing the answer to the first hop of a multi-hop question, our approach still outperforms it significantly.

In terms of unrelated knowledge stability, StableKE achieved accuracy scores of $87.03\%$ and $83.81\%$, significantly outperforming other methods. These results strongly suggest that our approach excels in preserving non-editable knowledge compared to alternative methods.

We also test the general ability of LLMs after knowledge editing. As depicted in Table~\ref{tab:main}, the MMLU performance of all methods decreases to varying degrees compared to the vanilla Vicuna model. However, StableKE stands out with a performance of $42.49\%$, significantly surpassing other methods. Furthermore, Table~\ref{tab:main} highlights that within the 'Multi-Hop' metric, StableKE excels in Composed-2Hop Accuracy (CoT) compared to ROME and MEMIT, which experience a decrease in CoT. This observation suggests that StableKE effectively processes instructions for multi-hop explanations, maintaining the CoT and instruction-following capabilities of LLMs.

Table \ref{tab:main} highlights the efficient performance of both StableKE and MEMIT in editing 1,000 knowledge triples simultaneously. However, in Appendix Table~\ref{tab:full_com} presents their performance when these 1,000 knowledge triples are divided into multiple editing sessions. It becomes apparent that both StableKE and MEMIT achieve optimal results when all knowledge edits are executed simultaneously. Nevertheless, as the $N_{seq}$ increases, there is a gradual decline in model stability across four different aspects. Notably, under the configuration of $N_{batch}=500$ and $N_{seq}=2$ the MEMIT method leads to model collapse.

\begin{figure*}
\centering
\includegraphics[width=1\linewidth]{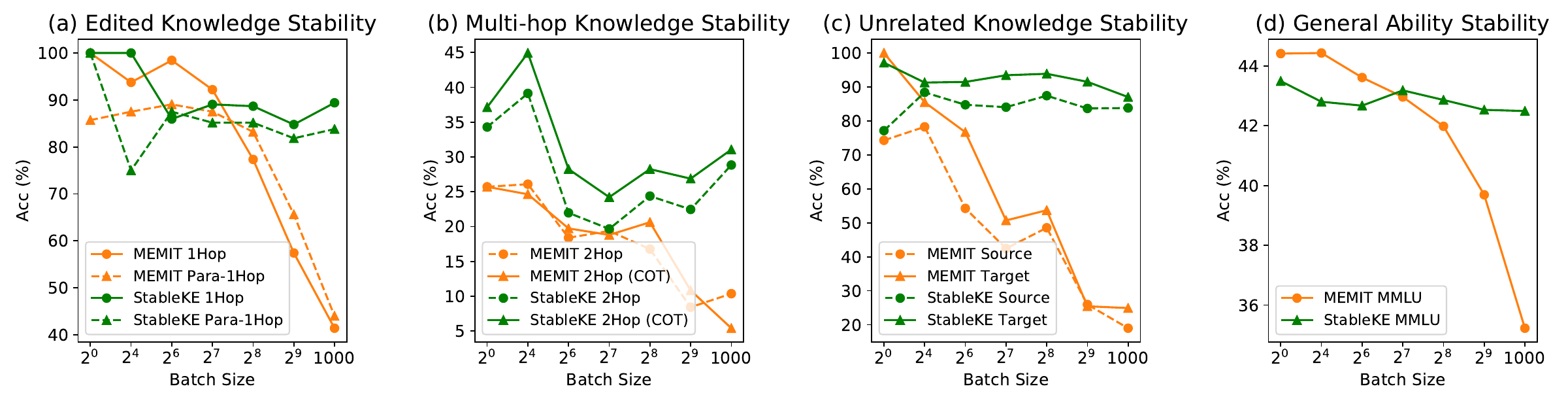}
\caption{Impact of batch size $N_{batch}$ on MEMIT and StableKE performance across four stability aspects.}
\label{fig:batch}
\vspace{-0.5cm}
\end{figure*}

\subsection{Batch Editing and Sequential Editing Analysis}

We evaluate the stability of knowledge editing methods under two distinct large-scale settings: batch editing and sequential editing to address RQ2. 

In the batch editing setup, in order to control variables, without loss of generality, we employ an exponential growth pattern of $N_{batch}=2^n$ to select data for editing, the performances are averaged over multiple runs. As depicted in Figure~\ref{fig:batch}, both our method and MEMIT demonstrate stability under the edited metric up to $N_{batch}=2^7$. However, beyond $2^7$ up to 1000 edited samples, the performance of MEMIT declines rapidly, whereas StableKE maintains stability. In terms of the multi-hop metric, StableKE consistently outperforms the MEMIT model across all stages, indicating its superior ability to facilitate the model in learning associations between pieces of knowledge. Furthermore, regarding the unrelated metric, StableKE's performance remains stable with an increasing number of edits, while MEMIT's performance experiences a rapid decline between $2^7$ and 1000 edited samples. Further details can be found in Table \ref{tab:full_batch}.
Moreover, the CoT performance of StableKE surpasses the original multi-hop QA in all training settings, indicating its superiority in preserving the instruction-following ability of the model and CoT capabilities. On the other hand, the CoT performance of MEMIT does not exhibit a significant difference from the standard multi-hop performance, suggesting that our method is more effective in maintaining these crucial capabilities.

\begin{figure*}
\centering
\includegraphics[width=1\linewidth]{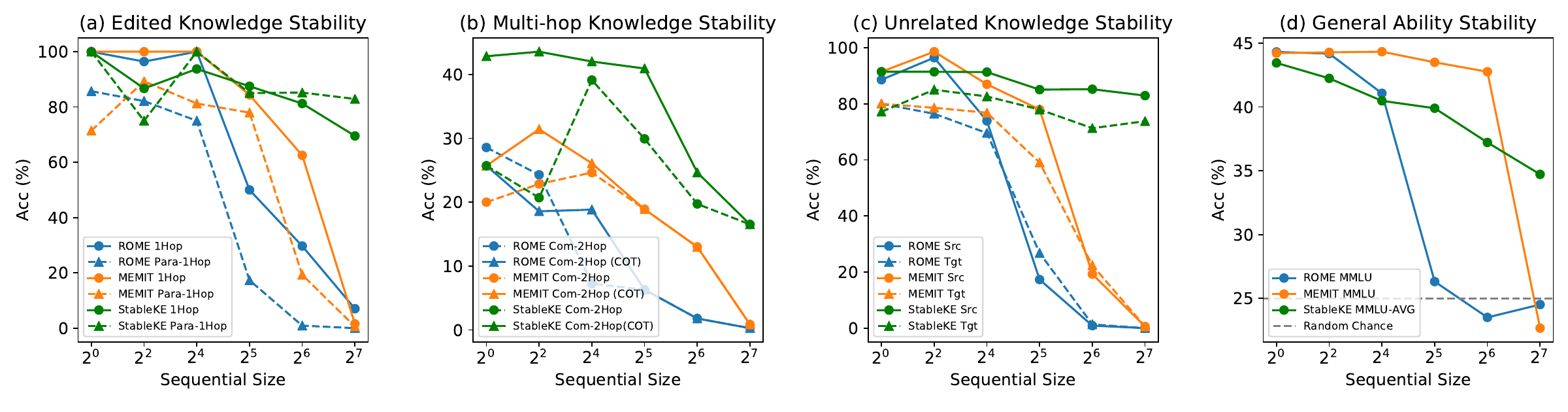}
\caption{Impact of sequential size $N_{seq}$ on MEMIT and StableKE performance across four stability aspects.}
\label{fig:seq}
\vspace{-0.5cm}
\end{figure*}

For sequential editing, in order to control variables, without loss of generality, we follow a criterion for data selection that adheres to an exponential growth paradigm, denoted by $N_{seq}=2^n$. However, as indicated in Figure~\ref{fig:seq}, when the dataset reaches $2^7$, both ROME and MEMIT experience model collapse due to the editing process. Consequently, we set the maximum value of $n$ to $7$, corresponding to a data volume of 128. Prior to reaching $2^4$, StableKE, MEMIT, and ROME all maintain stability across all four metrics. However, between $2^4$ and $2^7$ knowledge samples, the performance of ROME and MEMIT deteriorates rapidly, whereas our method remains stable. For more comprehensive details, please refer to the Appendix Table~\ref{tab:full_seq}.
Besides, the Com-2Hop CoT performance of StableKE consistently surpasses the original 2hop question-answering across all data points. On the other hand, MEMIT and ROME do not exhibit significant differences from multi-hop question-answering in terms of CoT performance. This underscores the advantage of our method in preserving the ability of the model to follow instructions and its multi-hop reasoning capabilities. As indicated in Table~\ref{tab:mmlu_seq}, after knowledge editing, the MMLU results of different methods all demonstrate varying degrees of decline compared to the original Vicuna model. Notably, ROME and MEMIT score $24.52\%$ and $22.67\%$ respectively, falling below $25\%$, suggesting that the models are nearly randomly answering questions. In contrast, StableKE's $34.72\%$ indicates that it better preserves the general performance of the model. 

\subsection{Impact of Semantic Paraphrase Quantity on Model Editing Performance}
\begin{figure}[h]
\centering
\includegraphics[width=0.9\linewidth]{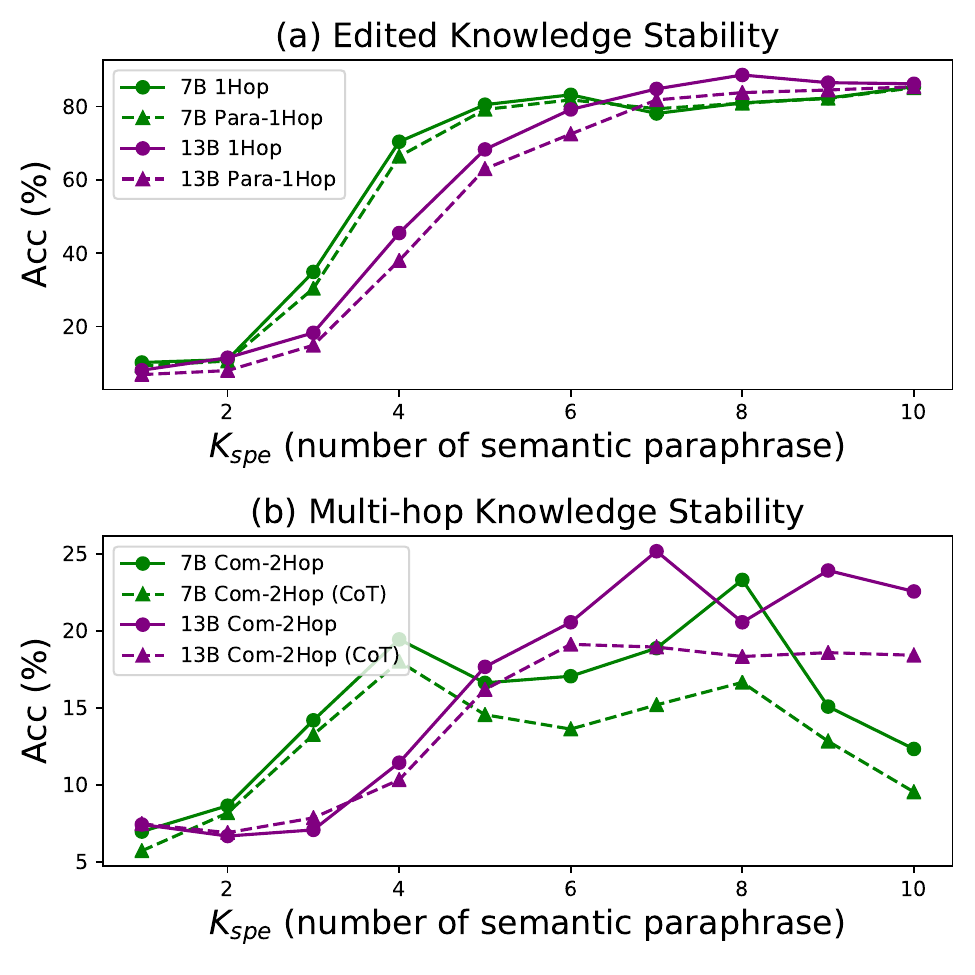}
\caption{Impact of semantic paraphrase quantity on StableKE performance in Vicuna-7b and Vicuna-13B.}
\label{fig:SPE}
\vspace{-0.5cm}
\end{figure}

We exam the performance of Vicuna-7B and Vicuna-13B models in response to variations in the number of semantic paraphrases. As depicted in Figure \ref{fig:SPE}, when the quantity of semantic paraphrases $K_{spe}$ is fewer than 2, 7B model exhibited lower accuracy in both edited knowledge stability and multi-hop knowledge stability. However, as $K_{spe}$ increase from 2 to 5, there is a significant and rapid improvement in the accuracy of 7B model ultimately reaching a plateau.

In contrast, 13B model demonstrated a slight lag in reaching this accuracy improvement plateau, yet it surpasses 7B model in terms of edited knowledge accuracy in its stable phase. Despite multi-hop knowledge not being a primary focus during the training process, the performance of 13B model in this area remained consistent with changes in the number of semantic paraphrases. Conversely, 7B model exhibited a decline, potentially indicating overfitting issues in edited knowledge. These findings suggest that increasing the model size necessitates incorporating greater diversity in knowledge editing, emphasizing significant change points and underscoring the importance of knowledge enhancement in enhancing the comprehension abilities of large language models. For detailed results, please refer to the Appendix Table~\ref{tab:7_SPE} and Table~\ref{tab:13_SPE}.

\subsection{Evaluating the Performance of StableKE Across Various Finetuning Methods}

This section addresses RQ4, focusing on the resource-demanding process of fine-tuning. We introduce LoRA~\cite{DBLP:conf/iclr/HuSWALWWC22}, a parameter-efficient finetuning technique commonly used in practice, to finetune language models in our StableKE method. As illustrated in Table~\ref{tab:main}, LoRA surpasses the performance of traditional approaches in the domain of knowledge editing, achieving significant results with limited parameter training.

Our method is designed without restrictions on the fine-tuning method, making it theoretically adaptable to various fine-tuning methods. We utilized the ChatGPT fine-tune API for knowledge editing purposes on ChatGPT 3.5, one of the most sophisticated closed-source models currently available. Considering costs, we randomly selected 16 factual triples from our dataset and successfully conducted knowledge editing on ChatGPT, as depicted in Table~\ref{tab:ChatGPT}. This demonstrates the efficacy of our approach for knowledge editing tasks on closed-source LLMs with available fine-tuning APIs.

\begin{table}
\centering
\resizebox{1\linewidth}{!}{
\begin{tabular}{ccccc}
\toprule
\multirow{2}{*}{\textbf{Method}} & \multicolumn{2}{c}{\textbf{Edited}} & \multicolumn{2}{c}{\textbf{Multi-Hop}}\\
\cmidrule(r){2-3} \cmidrule(r){4-5}
& \textbf{\small{1Hop-Acc}} & \textbf{\small{Para-1Hop-Acc}} & \textbf{\small{Com-2Hop-Acc}} & \textbf{\small{Com-2Hop-Acc (CoT)}}\\
\midrule
Finetune & 31.25 & 6.25 & 8.64 & 2.47\\
StableKE & 100.00 & 93.75 & 32.10 & 22.22\\
\bottomrule
\end{tabular}}
\caption{Performance of knowledge editing on ChatGPT using pure Finetune and StableKE.}
\label{tab:ChatGPT}
\vspace{-0.5cm}
\end{table}

\section{Conclusion}

In this study, we observed that most previous knowledge editing methods heavily rely on the assumption that knowledge is localized and isolated, leading to instability in knowledge editing methods. To verify these issues, we first developed a new knowledge editing benchmark, KEBench, which evaluates knowledge editing methods across four dimensions: edited knowledge stability, multi-hop knowledge stability, unrelated knowledge stability, and general ability stability.
Based on these findings, we introduced the StableKE method, which leverages knowledge augmentation rather than focusing solely on knowledge localization. We found that StableKE is a simple yet effective method for editing knowledge. The outstanding performances across all four stabilities demonstrate that the quality of the data is more crucial than the structure itself. This highlights the importance of paying close attention to data quality in this field.

\section*{Limitations}
Despite StableKE exhibiting stability across various knowledge editing setting, its overall performance exhibits a downward trend in a sequential editing setting as the number of editing operations increases. Moreover, while StableKE outperforms other knowledge editing methods in processing multi-hop knowledge, its accuracy in Com-2Hop significantly lags behind that in Decom-2Hop. This highlights a considerable opportunity for enhancing reasoning capabilities of the model.

\section*{Ethics Statement}
In this study, we conducted a thorough analysis of the knowledge editing process in LLMs and established four stability aspects to comprehensively evaluate the consistency and accuracy of knowledge editing methods. Through our proposed StableKE methods, we can effectively reduce the risks associated with misinformation, significantly enhancing stability of the edited model. Additionally, this study explores the stability of knowledge editing methods from four aspects, contributing to the development of more stable knowledge editing methods.

\bibliography{anthology,custom}

\appendix

\appendix

\section{Sampling Fact Chains from Wikidata}
\label{app:Sampling}
First, we eliminate all entities with different Q IDs but identical entity labels. Then, we filter out any first-hop and second-hop knowledge that forms circular relationships. Lastly, we discard factual triples if their first hop was edited and they recur in the second hop.

\section{Implementing Details}
\label{sec:implementation}
\textbf{Prompt} The edited knowledge triple $A(t_1^*)$ is utilized as the context specifically to finalize the knowledge editing process within the LLMs. 

\noindent\textbf{SERAC} We utilized the SERAC implementation provided by \cite{DBLP:journals/corr/abs-2308-07269} on the eazyedit platform. For the scope classifier, distilbert-base-cased was selected, while the counterfactual model employed was the llama-160m, as provided by \citet{miao2023specinfer}. During the training phase, we set the batch size to 100. All other training and inference parameters were kept at their default settings. For detailed configurations, please refer to the respective sources. 

\noindent\textbf{MEND} We employed the MEND implementation provided by \cite{DBLP:journals/corr/abs-2308-07269} via the eazyedit platform. During the training phase, we set the batch size to 500. This decision was informed by previous research, which reported suboptimal performance in the sequential edit of the MEND. Therefore, we opted for the largest feasible batch size within our resource constraints to minimize the number of sequential edits. All other training and inference parameters were retained at their default settings, as per the original model specifications.

\noindent\textbf{ROME} We implemented the ROME model using the framework provided by \cite{DBLP:journals/corr/abs-2308-07269} through eazyedit. Given that ROME permits only a single data edit per iteration, we report in table \ref{tab:main} the results obtained from a thousand successive edits using the same ROME model. All other training and inference parameters were maintained at their default settings.

\noindent\textbf{MEMIT} we utilized the MEMIT implementation provided by \cite{DBLP:journals/corr/abs-2308-07269} through eazyedit. As indicated in our results reported in table \ref{tab:full_batch}, MEMIT shows optimal performance when editing a thousand data entries per iteration. Consequently, we set the batch size to 1000 in our training configuration. All other training and inference parameters were kept at their default settings.

\noindent\textbf{ChatGPT Finetune} We employed the Fine-tuning API provided by OpenAI\footnote{https://platform.openai.com/docs/api-reference}. The experiment involved two methods of fine-tuning: data augmentation fine-tuning and standard fine-tuning. Each method was conducted over ten rounds of fine-tuning, and throughout the process, unless otherwise specified, we adhered to the default parameters set by OpenAI.

\noindent\textbf{StableKE} we utilized the instruction tuning and LoRA~\cite{DBLP:conf/iclr/HuSWALWWC22} implementation provided by \cite{DBLP:journals/corr/abs-2306-05685} through FastChat. we set the $K_{spe}$ and $K_{cde}$ to 5. All other training and inference parameters were kept at their default settings.

\section{Implementation Details of Semantic Paraphrase Enhancement}
\label{app:c}
We prompt ChatGPT (gpt-3.5-turbo) to automatically generate the semantic paraphrase answer $A_i^p(t_1^*)$. The prompt we used is shown in Table~\ref{tab:gen_spe}.

\begin{table*}[ht]
    \centering
    \small
    \noindent\fbox{%
    \begin{minipage}{2.0\columnwidth} 
\tt 

User: \newline I have a sentence: 'Sunak is the Prime Minister of Britain.' Could you provide five different ways to express this sentence? I'm looking for variations that are suitable for a formal speech. Please avoid using slang and keep the expressions concise. \newline

Assistant: {\newline 1. Sunak currently holds the position of Prime Minister in the United Kingdom.
\newline 2. The United Kingdom is presently led by Prime Minister Sunak.
\newline 3. Prime Minister Sunak serves as the head of government in Britain.
\newline 4. In the capacity of Prime Minister, Sunak leads the British government.
\newline 5. The office of Prime Minister in Britain is occupied by Sunak.}
    \end{minipage}
}

\caption{Demonstrating the application of ChatGPT (\texttt{gpt-3.5-turbo}) in generating semantic paraphrase answers $A_i^p(t_1^*)$ based on the 1hop answer $A(t_1^*)$.}
\label{tab:gen_spe}
\end{table*}

\section{Implementation Details of Contextual Description Enrichment}
We prompt ChatGPT (gpt-3.5-turbo) to automatically generate the document $doc_i^p(o)$. The prompt we used is shown in Table~\ref{tab:gen_cec}.

\begin{table*}[ht]
    \centering
    \small
    \noindent\fbox{%
    \begin{minipage}{2.0\columnwidth} 
\tt 

User: \newline I have a paragraph: 'Joe Biden currently assumes the dual role of head of state and head of government for the United States of America, which is a North American country using English as its official language. Additionally, its capital city is Washington, D.C.' Could you provide five different ways to express this paragraph? I'm looking for variations that are suitable for a formal speech. Please avoid using slang and keep the expressions concise. \newline

Assistant: {\newline 1. At present, Joe Biden fulfills the responsibilities of both the President and the chief executive in the United States, an English-speaking nation located in North America, with Washington, D.C. as its capital.
\newline 2. Currently serving as both the head of state and the chief administrator, Joe Biden leads the United States of America, which is situated in North America and designates English as its primary language, with its governmental center in Washington, D.C.
\newline 3. In his capacity as the leader of the United States, Joe Biden holds the twin positions of head of state and head of government in this North American English-speaking country, with Washington, D.C. as its capital.
\newline 4. Joe Biden presently occupies the dual roles of the nation's chief executive and principal leader in the United States, a country in North America where English is the official language and Washington, D.C. serves as the capital.
\newline 5. Joe Biden is currently vested with the dual responsibilities of head of state and head of government in the United States of America, a country on the North American continent that adopts English as its official language and has Washington, D.C. as its capital city.}
    \end{minipage}
}

\caption{Demonstrating the application of ChatGPT (\texttt{gpt-3.5-turbo}) in generating the document $doc_i^p(o)$ based on Wikidata document $doc(o)$.}
\label{tab:gen_cec}
\end{table*}

\begin{table*}[t]
\centering
\begin{tabular}{l|cccc|c}
\toprule
\textbf{Method} & \textbf{Humanities} & \textbf{STEM} & \textbf{Social} \textbf{Sciences} & \textbf{Other} & \textbf{Average} \\
\midrule
Vicuna & 46.94 & 36.11 & 50.33 & 48.13 & 44.48 \\
Finetune & 43.00 & 33.96 & 47.81 & 44.26 & 41.44 \\
MEND & 24.85 & 22.16 & 22.12 & 23.79 & 23.14 \\
ROME & 24.18 & 22.74 & 22.60 & 23.85 & 23.30 \\
MEMIT & 35.23 & 31.09 & 37.80 & 38.12 & 35.23 \\
\midrule
StableKE+LoRA & 42.96 & 33.24 & 45.11 & 43.69 & 40.48 \\
StableKE & 44.55 & 34.97 & 49.25 & 44.58 & 42.49 \\
\bottomrule
\end{tabular}
\caption{Accuracy performance of models edited with StableKE and other knowledge editing methods on MMLU.}
\label{tab:mmlu_main}
\end{table*}

\label{app:chatgpt}

\begin{table*}[t]
\centering
\begin{tabular}{l|cc|cccc|c}
\toprule
\textbf{Method} & \textbf{$N_{Batch}$} & \textbf{$N_{seq}$} & \textbf{Humanities} & \textbf{STEM} & \textbf{Social Sciences} & \textbf{Other} & \textbf{Average} \\
\midrule
Vicuna & - & - & 46.94 & 36.11 & 50.33 & 48.13 & 44.48 \\
\midrule
MEMIT&50&20&24.34&22.22&22.05&23.88&23.05\\
StableKE&50&20&44.04&33.30&46.62&44.99&41.38\\
\midrule
MEMIT&100&10&23.11&22.07&22.01&23.75&22.70\\
StableKE&100&10&44.05&34.29&47.38&44.42&41.72\\
\midrule
MEMIT&200&5&23.92&23.76&23.68&26.23&24.39\\
StableKE&200&5&44.72&34.57&48.40&44.61&42.22\\
\midrule
MEMIT&500&2&24.38&23.83&23.80&24.02&23.99\\
StableKE&500&2&44.47&35.35&49.31&45.35&42.79\\
\midrule
MEMIT&1000&1&35.23 & 31.09 & 37.80 & 38.12 & 35.23 \\
StableKE&1000&1& 44.55 & 34.97 & 49.25 & 44.58 & 42.49 \\
\bottomrule
\end{tabular}
\caption{Accuracy performance of StableKE and MEMIT on MMLU with different Batch Size and sequential.}
\label{tab:mmlu_com}
\end{table*}

\begin{table*}[t]
\centering
\begin{tabular}{l|c|cccc|c}
 \toprule
\textbf{Method} & \textbf{$N_{Batch}$} & \textbf{Humanities} & \textbf{STEM} & \textbf{Social Sciences} & \textbf{Other} & \textbf{Average} \\
\midrule
Vicuna & - & 46.94 & 36.11 & 50.33 & 48.13 & 44.48 \\
\midrule
MEMIT & 1 & 46.56 & 36.33 & 50.26 & 47.96 & 44.41 \\
StableKE & 1 & 45.38 & 35.33 & 49.80 & 46.93 & 43.49 \\
\midrule
MEMIT & 2 & 46.89 & 35.90 & 50.10 & 47.96 & 44.31 \\
StableKE & 2 & 45.20 & 35.66 & 49.61 & 46.51 & 43.41 \\
\midrule
MEMIT & 4 & 46.72 & 35.64 & 50.22 & 47.85 & 44.19 \\
StableKE & 4 & 45.40 & 35.29 & 49.47 & 46.04 & 43.18 \\
\midrule
MEMIT & 8 & 47.17 & 35.45 & 50.33 & 47.20 & 44.09 \\
StableKE & 8 & 45.32 & 35.53 & 49.16 & 46.30 & 43.24 \\
\midrule
MEMIT & 16 & 47.02 & 36.17 & 50.37 & 47.74 & 44.43 \\
StableKE & 16 & 44.56 & 34.54 & 49.38 & 46.26 & 42.80 \\
\midrule
MEMIT & 32 & 46.33 & 35.40 & 49.69 & 47.24 & 43.76 \\
StableKE & 32 & 44.52 & 34.41 & 49.69 & 46.38 & 42.84 \\
\midrule
MEMIT & 64 & 45.64 & 35.54 & 50.11 & 46.67 & 43.61 \\
StableKE & 64 & 45.14 & 34.18 & 49.62 & 45.52 & 42.67 \\
\midrule
MEMIT & 128 & 44.56 & 35.41 & 49.38 & 45.81 & 42.96 \\
StableKE & 128 & 45.48 & 35.19 & 49.48 & 46.10 & 43.18 \\
\midrule
MEMIT & 256 & 43.08 & 34.04 & 48.15 & 45.95 & 41.98 \\
StableKE & 256 & 45.34 & 34.63 & 49.61 & 45.53 & 42.86 \\
\midrule
MEMIT & 512 & 40.63 & 33.52 & 44.48 & 42.70 & 39.69 \\
StableKE & 512 & 45.24 & 33.86 & 49.45 & 45.41 & 42.53 \\
\midrule
MEMIT & 1000 & 31.09 & 35.23 & 37.80 & 38.12 & 35.23 \\
StableKE & 1000 & 44.55 & 34.97 & 49.25 & 44.58 & 42.49 \\
\bottomrule
\end{tabular}
\caption{Accuracy performance of StableKE and MEMIT on MMLU with different Batch Size.}
\label{tab:mmlu_batch}
\end{table*}

\begin{table*}[t]
\centering
\begin{tabular}{l|c|cccc|c}
\toprule
\textbf{Method} & \textbf{$N_{seq}$} & \textbf{Humanities} & \textbf{STEM} & \textbf{Social Sciences} & \textbf{Other} & \textbf{Average}  \\
\midrule
Vicuna & - &46.94 & 36.11 & 50.33 & 48.13 & 44.48 \\
\midrule
ROME & 1 & 47.40 & 35.61 & 50.11 & 47.89 & 44.31\\
MEMIT & 1 & 46.40 & 36.48 & 49.56 & 47.74 & 44.22\\
StableKE & 1 & 45.37 & 35.28 & 49.80 & 46.89 & 43.45\\
\midrule
ROME & 2 & 46.91 & 35.59 & 49.70 & 47.55 & 44.03\\
MEMIT & 2 & 46.59 & 35.80 & 49.90 & 47.93 & 44.17\\
StableKE & 2 & 44.88 & 36.18 & 48.30 & 46.21 & 43.15\\
\midrule
ROME & 4 & 47.17 & 35.94 & 49.51 & 47.64 & 44.18\\
MEMIT & 4 & 47.29 & 35.64 & 50.12 & 47.80 & 44.28\\
StableKE & 4 & 44.59 & 34.73 & 47.35 & 45.47 & 42.24\\
\midrule
ROME & 8 & 45.44 & 35.87 & 48.36 & 47.64 & 43.54\\
MEMIT & 8 & 47.19 & 35.88 & 50.63 & 47.66 & 44.41\\
StableKE & 8 & 43.57 & 34.08 & 47.73 & 43.77 & 41.46\\
\midrule
ROME & 16 & 41.25 & 34.77 & 46.07 & 44.72 & 41.07\\
MEMIT & 16 & 47.46 & 35.64 & 50.47 & 47.54 & 44.33\\
StableKE & 16 & 43.25 & 33.33 & 46.47 & 42.18 & 40.48\\
\midrule
ROME & 32 & 24.75 & 27.82 & 25.13 & 26.75 & 26.32\\
MEMIT & 32 & 46.02 & 35.34 & 50.03 & 46.24 & 43.50\\
StableKE & 32 & 41.86 & 32.33 & 44.77 & 43.75 & 39.90\\
\midrule
ROME & 64 & 24.76 & 22.23 & 22.41 & 25.01 & 23.51\\
MEMIT & 64 & 44.53 & 35.80 & 48.70 & 45.11 & 42.76\\
StableKE & 64 & 39.08 & 31.43 & 40.88 & 39.92 & 37.22\\
\midrule
ROME & 128 & 24.93 & 23.10 & 23.71 & 26.69 & 24.52\\
MEMIT & 128 & 22.72 & 22.11 & 22.13 & 23.83 & 22.67\\
StableKE & 128 & 37.07 & 29.61 & 37.13 & 37.21 & 34.72\\
\bottomrule
\end{tabular}
\caption{Accuracy performance of StableKE and MEMIT on MMLU with different Sequential Size.}
\label{tab:mmlu_seq}
\end{table*}

\begin{table*}[t]
\centering
\resizebox{1\linewidth}{!}{
\begin{tabular}{c|ccccccc}
\toprule
\multirow{2}{*}{\textbf{$K_{SPE}$}} & \multicolumn{2}{c}{\textbf{Edited}} & \multicolumn{3}{c}{\textbf{Multi-Hop}} & \multicolumn{2}{c}{\textbf{Unrelated}}\\
\cmidrule(r){2-3} \cmidrule(r){4-6} \cmidrule(r){7-8}
 & \textbf{1Hop-Acc} & \textbf{Para-1Hop-Acc} & \textbf{Decom-2Hop-Acc} & \textbf{Com-2Hop-Acc} & \textbf{Com-2Hop-Acc (CoT)} & \textbf{Src-Acc} & \textbf{Tgt-Acc} \\ 
\midrule
1 & 10.20 & 9.30 & 1.89 & 5.72 & 6.97 & 26.48 & 24.23\\
2 & 11.00 & 10.60 & 5.47 & 8.18 & 8.65 & 50.36 & 41.82\\
3 & 34.90 & 30.40 & 16.51 & 13.26 & 14.19 & 55.58 & 47.43\\
4 & 70.40 & 66.40 & 39.42 & 18.01 & 19.44 & 62.47 & 57.83\\
5 & 80.50 & 79.20 & 44.57 & 14.55 & 16.62 & 56.15 & 53.68\\
6 & 78.10 & 79.40 & 53.97 & 15.19 & 18.87 & 75.27 & 65.05\\
7 & 81.00 & 80.90 & 60.01 & 16.65 & 23.30 & 77.56 & 70.44\\
8 & 82.30 & 82.20 & 53.15 & 12.83 & 15.08 & 70.51 & 61.87\\
9 & 85.40 & 85.10 & 57.33 & 9.54 & 12.33 & 68.33 & 64.26\\
10 & 86.20 & 86.00 & 57.40 & 18.19 & 19.62 & 74.62 & 64.22\\
\bottomrule
\end{tabular}}
\caption{Impact of the number of semantic paraphrases on edited knowledge stability, multi-hop knowledge stability, and unrelated knowledge stability of the Vicuna-7B model.}
\label{tab:7_SPE}
\end{table*}

\begin{table*}[t]
\centering
\resizebox{1\linewidth}{!}{
\begin{tabular}{c|ccccccc}
\toprule
\multirow{2}{*}{\textbf{$K_{SPE}$}} & \multicolumn{2}{c}{\textbf{Edited}} & \multicolumn{3}{c}{\textbf{Multi-Hop}} & \multicolumn{2}{c}{\textbf{Unrelated}} \\
\cmidrule(r){2-3} \cmidrule(r){4-6} \cmidrule(r){7-8}
 & \textbf{1Hop-Acc} & \textbf{Para-1Hop-Acc} & \textbf{Decom-2Hop-Acc} & \textbf{Com-2Hop-Acc} & \textbf{Com-2Hop-Acc (CoT)} & \textbf{Src-Acc} & \textbf{Tgt-Acc} \\ 
\midrule
1 & 8.10 & 6.90 & 2.54 & 7.47 & 7.43 & 38.21 & 35.49 \\
2 & 11.50 & 8.00 & 3.50 & 6.90 & 6.68 & 40.46 & 39.42 \\
3 & 18.30 & 14.90 & 9.47 & 7.86 & 7.08 & 48.25 & 50.07 \\
4 & 45.50 & 38.00 & 24.80 & 10.33 & 11.44 & 63.33 & 62.29 \\
5 & 68.30 & 63.00 & 41.57 & 16.19 & 17.66 & 67.16 & 61.44 \\
6 & 79.20 & 72.50 & 51.18 & 19.12 & 20.55 & 66.40 & 63.69 \\
7 & 84.80 & 81.80 & 58.47 & 18.94 & 25.16 & 79.31 & 67.58 \\
8 & 88.60 & 83.80 & 59.26 & 18.33 & 20.55 & 70.44 & 66.30 \\
9 & 86.50 & 84.50 & 63.44 & 18.58 & 23.91 & 78.91 & 71.37 \\
10 & 87.60 & 85.40 & 60.19 & 20.30 & 26.16 & 72.59 & 67.12 \\
\bottomrule
\end{tabular}}
\caption{Impact of the number of semantic paraphrases on edited knowledge stability, multi-hop knowledge stability, and unrelated knowledge stability of the Vicuna-13B model.}
\label{tab:13_SPE}
\end{table*}

\begin{table*}[t]
\centering
\resizebox{1\linewidth}{!}{
\begin{tabular}{c|cc|cccccccc}
\toprule
\multirow{2}{*}{\textbf{Method}} & \multirow{2}{*}{\textbf{$N_{Batch}$}} & \multirow{2}{*}{\textbf{$N_{seq}$}} & \multicolumn{2}{c}{\textbf{Edited}} & \multicolumn{3}{c}{\textbf{Multi-Hop}} & \multicolumn{2}{c}{\textbf{Unrelated}} & \multicolumn{1}{c}{\textbf{General}} \\
\cmidrule(r){4-5} \cmidrule(r){6-8} \cmidrule(r){9-10}\cmidrule(r){11-11}
 & & & \textbf{1Hop-Acc} & \textbf{Para-1Hop-Acc} & \textbf{Decom-2Hop-Acc} & \textbf{Com-2Hop-Acc} & \textbf{Com-2Hop-Acc (CoT)} & \textbf{Src-Acc} & \textbf{Tgt-Acc} & \textbf{MMLU}\\ 
\midrule
MEMIT & 50 & 20 & 0.10& 0.00& 0.25& 0.21& 0.21 & 1.93& 3.00 &23.05\\
StableKE & 50 & 20 & 89.90& 85.60& 73.98& 20.44& 24.87 & 85.31& 82.31 & 41.38\\
\midrule
MEMIT & 100 & 10 & 0.10& 0.00& 0.04& 0.18& 0.11 & 0.29& 0.86 & 22.70\\
StableKE & 100 & 10 & 87.30& 87.20& 73.95& 20.44& 25.38 & 81.45& 81.81 & 41.72\\
\midrule
MEMIT & 200 & 5 & 0.10& 0.10& 0.00& 0.04& 0.04 & 0.00& 0.18 & 24.39\\
StableKE & 200 & 5 & 89.30& 86.90& 74.23& 25.09& 26.80 & 80.02& 81.88 & 42.22\\
\midrule
MEMIT & 500 & 2 & 9.90& 11.30& 0.25& 1.39& 1.36 & 1.93& 3.00 & 23.99\\
StableKE & 500 & 2 & 88.90& 81.60& 75.52& 28.73& 31.09 & 88.13& 83.38 & 42.79\\
\midrule
MEMIT & 1000 & 1 & 41.40& 44.00& 13.27& 10.36& 5.40 & 24.91& 18.98 & 40.48\\
StableKE & 1000 & 1 & 89.40& 83.80& 77.09& 28.84& 31.06 & 87.03& 83.81 & 42.49\\
\bottomrule
\end{tabular}}
\caption{Performance of StableKE and MEMIT on four stability metrics under different batch size and sequential size settings.}
\label{tab:full_com}
\end{table*}

\begin{table*}
\centering
\resizebox{1\linewidth}{!}{
\begin{tabular}{c|c|cccccccc}
\toprule
\multirow{2}{*}{\textbf{Method}} & \multirow{2}{*}{\textbf{$N_{Batch}$}} & \multicolumn{2}{c}{\textbf{Edited}} & \multicolumn{3}{c}{\textbf{Multi-Hop}} & \multicolumn{2}{c}{\textbf{Unrelated}} & \multicolumn{1}{c}{\textbf{General}} \\
\cmidrule(r){3-4} \cmidrule(r){5-7} \cmidrule(r){8-9} \cmidrule(r){10-10}
 & & \multicolumn{1}{c}{\textbf{\small{1Hop-Acc}}} & \multicolumn{1}{c}{\textbf{\small{Para-1Hop-Acc}}} & \multicolumn{1}{c}{\textbf{\small{Decom-2Hop-Acc}}} & \multicolumn{1}{c}{\textbf{\small{Com-2Hop-Acc}}} & \multicolumn{1}{c}{\textbf{\small{Com-2Hop-Acc (CoT)}}} & \multicolumn{1}{c}{\textbf{\small{Src-Acc}}} & \multicolumn{1}{c}{\textbf{\small{Tgt-Acc}}} & \multicolumn{1}{c}{\textbf{\small{MMLU}}} \\ 
\midrule
MEMIT & 1 & 100.00 & 85.71 & 74.29 & 25.71 & 25.71 & 100.00 & 74.29 &44.41 \\
StableKE & 1 & 100.00 & 100.00 & 74.29 & 34.29 & 37.14 & 97.14 & 77.14 &43.49 \\
\midrule
MEMIT & 2 & 85.71 & 78.57 & 68.57 & 25.71 & 24.29 & 85.71 & 68.57 & 44.31\\
StableKE & 2 & 85.71 & 78.57 & 71.43 & 28.57 & 47.14 & 98.57 & 91.43 & 43.41 \\
\midrule
MEMIT & 4 & 78.57 & 71.43 & 67.14 & 25.56 & 25.89 & 85.71 & 69.29 & 44.19 \\
StableKE & 4 & 85.71 & 82.14 & 69.29 & 32.14 & 47.14 & 92.86 & 85.71 & 43.18 \\
\midrule
MEMIT & 8 & 82.14 & 80.36 & 63.71 & 20.72 & 21.92 & 80.14 & 65.95 & 44.09 \\
StableKE & 8 & 80.36 & 78.57 & 70.63 & 34.05 & 47.01 & 93.06 & 86.02 & 43.24 \\
\midrule
MEMIT & 16 & 93.75& 87.50& 75.36& 26.09& 24.64 & 85.51& 78.26&44.43\\
StableKE & 16 & 100.00& 75.00& 75.36& 39.13& 44.93 & 91.30& 88.41&42.80\\
\midrule
MEMIT & 32 & 93.75& 87.50& 65.35& 19.69& 18.90 & 81.89& 66.14&43.76\\
StableKE & 32 & 90.63& 81.25& 77.95& 29.13& 43.31 & 93.70& 84.25&42.84\\
\midrule
MEMIT & 64 & 98.44& 89.06& 52.02& 18.39& 19.73 & 76.68& 54.26&43.61\\
StableKE & 64 & 85.94& 87.50& 78.92& 21.97& 28.25 & 91.48& 84.75&42.67\\
\midrule
MEMIT & 128 & 92.19& 87.50& 39.60& 19.37& 18.80 & 50.71& 42.45&42.96\\
StableKE & 128 & 89.06& 85.16& 78.63& 19.66& 24.22 & 93.45& 84.05&43.18\\
\midrule
MEMIT & 256 & 77.34& 83.20& 40.71& 16.76& 20.62 & 53.71& 48.55&41.98\\
StableKE & 256 & 88.67& 85.16& 79.16& 24.38& 28.25 & 93.88& 87.43&42.86\\
\midrule
MEMIT & 512 & 57.42& 65.63& 16.16& 8.42& 10.83 & 25.44& 25.96&39.69\\
StableKE & 512 & 84.77& 81.84& 74.38& 22.46& 26.88 & 91.52& 83.67&42.53\\
\midrule
MEMIT & 1000 & 41.40 & 44.00 & 13.37 & 10.36 & 5.40 & 24.91 & 18.98&35.23 \\
StableKE & 1000 & 89.40& 83.80& 77.09& 28.84& 31.06 & 87.03& 83.81&42.49\\
\bottomrule
\end{tabular}}
\caption{Performance of StableKE, MEMIT on four stability metrics under different batch size settings.}
\label{tab:full_batch}
\end{table*}

\begin{table*}[t]
\centering
\resizebox{1\linewidth}{!}{
\begin{tabular}{c|c|cccccccc}
\toprule
\multirow{2}{*}{\textbf{Method}} & \multirow{2}{*}{\textbf{$N_{seq}$}} & \multicolumn{2}{c}{\textbf{Edited}} & \multicolumn{3}{c}{\textbf{Multi-Hop}} & \multicolumn{2}{c}{\textbf{Unrelated}} & \multicolumn{1}{c}{\textbf{General}}\\
\cmidrule(r){3-4} \cmidrule(r){5-7} \cmidrule(r){8-9} \cmidrule(r){10-10}
 & & \multicolumn{1}{c}{\textbf{\small{1Hop-Acc}}} & \multicolumn{1}{c}{\textbf{\small{Para-1Hop-Acc}}} & \multicolumn{1}{c}{\textbf{\small{Decom-2Hop-Acc}}} & \multicolumn{1}{c}{\textbf{\small{Com-2Hop-Acc}}} & \multicolumn{1}{c}{\textbf{\small{Com-2Hop-Acc (CoT)}}} & \multicolumn{1}{c}{\textbf{\small{Src-Acc}}} & \multicolumn{1}{c}{\textbf{\small{Tgt-Acc}}} & \multicolumn{1}{c}{\textbf{\small{MMLU}}} \\ 
\midrule
ROME & 1 & 100.00 & 85.71 & 88.57 & 80.00 & 80.00 & 28.57 & 25.71 & 44.31 \\
MEMIT & 1 & 100.00 & 71.43 & 91.43 & 80.00 & 80.00 & 20.00 & 25.71 & 44.22 \\
StableKE & 1 & 100.00 & 100.00 & 91.43 & 77.14 & 68.57 & 25.71 & 42.86 & 43.45 \\
\midrule
ROME & 2 & 100.00 & 92.86 & 95.71 & 82.86 & 80.00 & 25.71 & 30.00 & 44.03 \\
MEMIT & 2 & 100.00 & 85.71 & 95.71 & 77.14 & 77.14 & 27.14 & 27.14 & 44.17 \\
StableKE & 2 & 100.00 & 100.00 & 85.70 & 75.70 & 61.40 & 45.70 & 51.40 & 43.15 \\
\midrule
ROME & 4 & 96.43 & 82.14 & 96.43 & 76.43 & 74.29 & 24.29 & 18.57 & 44.18 \\
MEMIT & 4 & 100.00 & 89.29 & 98.57 & 78.57 & 78.57 & 22.86 & 31.43 & 44.28 \\
StableKE & 4 & 86.70 & 75.00 & 91.40 & 85.00 & 67.90 & 20.70 & 43.57 & 42.24 \\
\midrule
ROME & 8 & 96.43 & 89.29 & 92.32 & 72.51 & 72.15 & 22.02 & 20.14 & 43.54 \\
MEMIT & 8 & 91.07 & 91.07 & 95.28 & 74.18 & 73.46 & 32.65 & 33.47 & 44.41 \\
StableKE & 8 & 89.29 & 85.71 & 84.83 & 84.66 & 70.33 & 28.61 & 32.40 & 41.46 \\
\midrule
ROME & 16 & 100.00& 75.00& 66.67& 7.25& 18.84 & 73.91& 69.57&41.07\\
MEMIT & 16 & 93.75& 100.00& 76.81& 39.13& 42.0 & 91.30& 82.61&44.33\\
StableKE & 16 & 93.75& 100.00& 76.81& 39.13& 42.03 & 91.30& 82.4&40.48\\
\midrule
ROME & 32 & 53.13& 50.00& 14.17& 3.94& 6.30 & 17.32& 26.77&26.32\\
MEMIT & 32 & 93.75& 84.38& 55.91& 21.26& 18.90 & 77.95& 59.06&43.50\\
StableKE & 32 & 93.75& 87.50& 66.93& 29.92& 40.94 & 85.04& 77.95&39.90\\
\midrule
ROME & 64 & 34.38& 29.69& 0.00& 1.79& 1.79 & 0.90& 1.35&23.51\\
MEMIT & 64 & 57.81& 62.50& 14.80& 13.90& 13.00 & 19.28& 22.42&42.76\\
StableKE & 64 & 81.25& 81.25& 64.57& 19.73& 24.66 & 85.20& 71.30&37.22\\
\midrule
ROME & 128 & 7.03& 7.03& 0.00& 0.28& 0.28 & 0.00& 0.00&24.52\\
MEMIT & 128 & 1.56& 1.56& 0.00& 0.85& 0.85 & 0.57& 0.57&22.67\\
StableKE & 128 & 73.44& 69.53& 55.56& 15.95& 16.52 & 82.91& 73.79&34.72\\
\bottomrule
\end{tabular}}
\caption{Performance of StableKE, MEMIT and ROME on four stability metrics under different sequential size settings.}
\label{tab:full_seq}
\end{table*}

\clearpage

\end{document}